\documentclass{article}
\usepackage{iclr2027_conference,times}
\usepackage{amsmath,amssymb,booktabs,graphicx,tabularx,array}
\usepackage{algorithm}
\usepackage{algpseudocode}
\usepackage{flafter}
\usepackage{capt-of}
\usepackage{xcolor}
\usepackage{xspace}
\usepackage{hyperref}
\usepackage{url}
\hypersetup{colorlinks=true,linkcolor=black,citecolor=black,urlcolor=blue,
  pdftitle={ReAT: Residual Action Tokenization for Autoregressive Robot Policies},
  pdfauthor={Dijia Zhan, Zhuo Li, Xuemiao Xu, Sun Zheng, Jinyi Li, Wenjun Wang, Jie Tang}}
\newcommand{\method}{ReAT\xspace}
\definecolor{semgray}{gray}{0.45}
\newcommand{\sem}[1]{\mathbin{\textcolor{semgray}{\pm}}\textcolor{semgray}{#1}}

\newcommand{\R}{\mathbb{R}}
\newcommand{\E}{\mathbb{E}}
\newcommand{\norm}[1]{\left\lVert#1\right\rVert}

\title{ReAT: Residual Action Tokenization\\for Autoregressive Robot Policies}
\author{\textbf{Dijia Zhan$^{1*}$ \& Zhuo Li$^{2*}$ \& Xuemiao Xu$^1$ \& Sun Zheng$^2$} \\
\textbf{Jinyi Li$^1$ \& Wenjun Wang$^3$ \& Jie Tang$^1$} \\
$^1$South China University of Technology (scut.edu.cn) \\
$^2$The Chinese University of Hong Kong (cuhk.edu.hk) \\
$^3$Hong Kong Polytechnic University (polyu.edu.hk) \quad $^*$Equal contribution.}
\iclrfinalcopy
\begin{document}
\maketitle
\fancyhead{}

\begin{abstract}
Autoregressive robot policies rely on action tokenizers to convert continuous action chunks into compact sequences of discrete prediction targets. Existing tokenizers can reconstruct action chunks from discrete codes, but often leave implicit what each individual token contributes to the reconstructed action. We introduce Residual Action Tokenization (\method), which gives each token an explicit role in refining the current action reconstruction. It uses the decoded reconstruction as feedback to construct the next token from the remaining action error. This closes the loop between encoding and decoding, making each new token responsive to the errors left by the actual decoded prefix. Across 19 manipulation tasks spanning three simulation benchmarks and real-world settings, 8-token \method policies outperform the evaluated baselines in task success; on LIBERO, even 4-token \method surpasses the evaluated 8-token action-tokenization baselines at half the token budget. These results suggest a path toward action tokenization that supports self-corrective policy learning, helping autoregressive policies learn to progressively refine their own action plans.
\end{abstract}

\begin{figure}[!h]
  \centering
  \includegraphics[width=\linewidth]{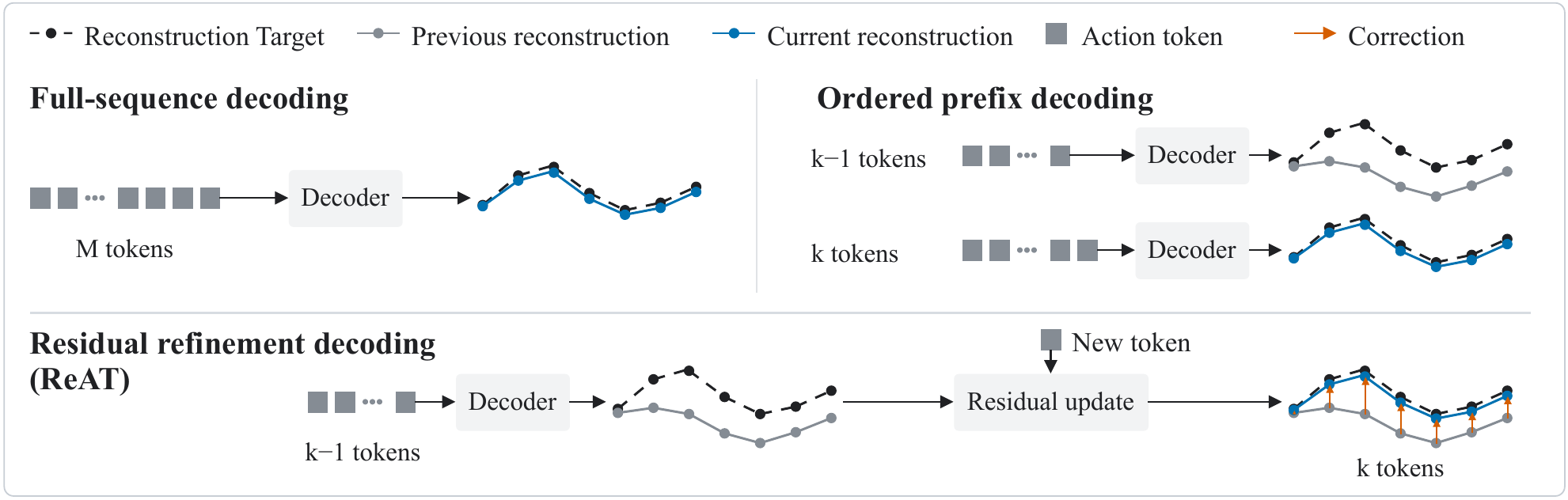}
  \caption{\textbf{Comparison of action-token decoding strategies.} Full-sequence decoding requires all $M$ tokens, while ordered prefix decoding reconstructs a complete action chunk from any nonempty prefix. \method combines the existing draft with each new token to produce a learned correction. Orange arrows indicate the corrections applied to the previous reconstruction in \method; target trajectories are shown only for reference.}
  \label{fig:overview}
\end{figure}

\section{Introduction}
\begin{figure}[!t]
  \centering
  \includegraphics[width=\linewidth]{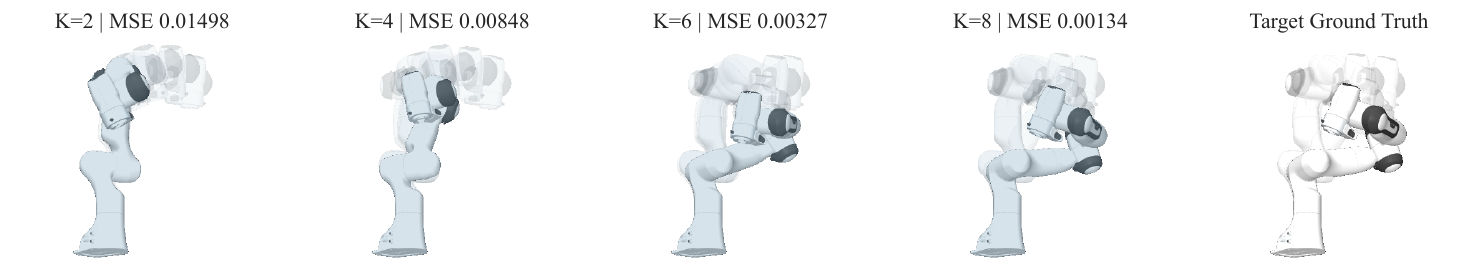}
  \caption{\textbf{Arm trajectory reconstruction with residual tokens.} A 32-step LIBERO example at $K=2,4,6,8$ using an earlier \method tokenizer. Five sampled poses show translucent motion history and an opaque final pose. MSE is computed over all steps and normalized action dimensions of this example.}
  \label{fig:libero-arms}
\end{figure}

Autoregressive robot policies generate future actions by predicting discrete tokens one at a time~\citep{brohan2023rt2,kim2024openvla}. To connect this prediction process to continuous motion, action tokenizers encode future action chunks into token sequences and decode the predicted tokens back into executable actions~\citep{mete2024quest,pertsch2025fast}. The encoded sequences provide the policy's training targets, so the tokenizer determines how action information is distributed across tokens and how each token depends on its predecessors~\citep{liu2026oat,dong2026actioncodec}. The question is then how to structure these targets so that each token has a clear role in reconstructing the action chunk.

Learned action tokenizers use reconstruction objectives to compress action chunks into discrete sequences for policy learning~\citep{lee2024vqbet,mete2024quest}. Recent work further shows that reconstruction fidelity alone is insufficient to characterize these representations, as token organization also affects downstream policy learning~\citep{dong2026actioncodec}. Ordering constraints and prefix supervision provide additional structure by encouraging earlier tokens to preserve essential action information and later tokens to add detail~\citep{liu2026oat}. However, these objectives do not by themselves explicitly define each token's contribution relative to the actions already reconstructed. That contribution can remain implicit in the learned encoder--decoder mapping, leaving an opportunity to ground successive tokens in the action information still missing from the current reconstruction.

The residual between a target action chunk and its current reconstruction provides a concrete description of this missing information. Because it is measured in the reconstructed actions, it reflects what preceding tokens have actually recovered, including the effects of quantization and decoding. Encoding this residual gives the next token an explicit target tied to the reconstruction produced so far. This motivates Residual Action Tokenization (\method), which structures action sequences as successive corrections to an evolving action draft.

\method implements this formulation through explicit reconstruction feedback. During tokenizer training, the encoder maps the remaining reconstruction error to a discrete token, and the decoder uses that token together with the current action draft to produce a learned correction (Figure~\ref{fig:overview}). The updated draft then provides the basis for constructing the next token. Each additional token thus represents a correction to the existing action draft, with Figure~\ref{fig:libero-arms} illustrating how reconstruction improves as tokens are added. The resulting residual tokens provide structured prediction targets for observation-conditioned autoregressive policies.

We evaluate \method on 19 manipulation tasks spanning LIBERO, MetaWorld, RoboCasa, and real-world bucket stacking. With 8 tokens, \method outperforms the evaluated action-tokenization and diffusion-policy baselines in task success on all three simulation benchmarks and the real-world task. On LIBERO, it surpasses the evaluated 8-token action-tokenization baselines using only 4 tokens. LIBERO ablations further support the contribution of explicit reconstruction feedback to policy performance and show that prefix supervision enables effective control with shorter token sequences.

In summary, our contributions are:
\begin{itemize}
  \item We introduce \method, an action tokenizer that constructs each successive token from the residual left by the decoded prefix, explicitly incorporating quantization and decoding effects into token construction.
  \item We combine a draft-conditioned recurrent decoder with prefix reconstruction supervision to turn residual tokens into successive action corrections. Each nonempty prefix reconstructs a complete action chunk, allowing a single autoregressive policy to use different token budgets without retraining.
  \item Experiments across three simulation benchmarks and real-world bucket stacking show that 8-token \method outperforms the evaluated action-tokenization and diffusion-policy baselines in task success. On LIBERO, 4-token \method also surpasses the evaluated 8-token action-tokenization baselines, while mechanism ablations support the contributions of reconstruction feedback and prefix supervision.
\end{itemize}

\section{Related Work}
\paragraph{Generative robot policies.}
Robot policies generate actions through continuous representations or discrete token sequences. Continuous approaches include variational action-chunk prediction~\citep{zhao2023act}, diffusion-based generation~\citep{chi2023diffusion}, and flow-matching action experts~\citep{black2024pi0,nvidia2025groot}. Discrete autoregressive policies predict action symbols sequentially~\citep{kim2024openvla}, while recent extensions explore parallel continuous-chunk decoding~\citep{kim2025oft} and discrete-action pretraining followed by flow-matching post-training~\citep{physicalintelligence2025pi05}. HiFlow further combines scale-wise autoregressive generation with flow matching in continuous action space~\citep{yashima2026hiflow}. These approaches primarily advance the mechanisms for generating actions, while leaving open how discrete action targets should be structured to support progressive correction. \method addresses this representation question by making residual correction the organizing principle of action tokenization, providing autoregressive policies with corrective prediction targets.

\paragraph{Action tokenization and ordered prefixes.}
Learned action tokenizers compress action chunks into quantized representations for policy learning~\citep{lee2024vqbet,mete2024quest,wang2025vqvla}, while FAST exploits temporal regularity through frequency-domain coding and byte-pair encoding~\citep{pertsch2025fast}. Beyond compression, CARP organizes action representations into scales for next-scale prediction~\citep{gong2025carp}, and ActionCodec examines how token structure affects downstream policy optimization~\citep{dong2026actioncodec}. Ordered tokenizers use nested dropout to prioritize information in early prefixes~\citep{rippel2014nested}, enabling flexible-length reconstruction in image and action domains~\citep{bachmann2025flextok,liu2026oat}. Compact representations and ordered prefixes improve the usability of action tokens, but do not by themselves make the remaining action reconstruction error an explicit basis for subsequent token construction. \method introduces this connection, giving successive tokens a correction target defined by what earlier tokens have actually reconstructed.

\paragraph{Residual coding and reconstruction feedback.}
Residual quantization successively approximates latent features, with applications in image generation~\citep{lee2022rq}, neural audio compression~\citep{zeghidour2021soundstream,defossez2022encodec}, and action tokenization~\citep{lee2024vqbet,wang2025vqvla,gong2025carp}. However, latent approximation error does not directly characterize the error remaining in decoded actions. Recurrent image codecs address the analogous distinction by feeding decoded-output errors back into progressive compression~\citep{toderici2016compression,toderici2016fullresolution}. \method brings this reconstruction-feedback perspective to action tokenization, constructing corrective representations for autoregressive policy learning.

\section{Preliminaries}
\label{sec:preliminaries}
Let $o_t$ denote an observation containing camera images and proprioception, and let $c$ identify a task. A policy uses an observation history $O_t=o_{t-H_o+1:t}$ of length $H_o$ to predict a chunk of $H$ actions, each with $d$ control dimensions. The robot executes the first $H_e\leq H$ commands before observing and replanning.

A tokenizer maps an expert action chunk $A\in\R^{H\times d}$ to a token sequence $z_{1:M}$, where $M$ is the maximum token budget and $z_k\in\{0,\ldots,V-1\}$ indexes a vocabulary of size $V$. We seek a prefix-decodable representation: every nonempty prefix $z_{1:K}$, $1\leq K\leq M$, should decode to a complete reconstruction $\widehat A^{(K)}\in\R^{H\times d}$. The tokenizer supplies training labels for an autoregressive policy $\pi(z_k\mid O_t,c,z_{<k})$. At deployment, the policy predicts tokens from observations and the decoder converts them into actions, without access to the expert chunk.

\begin{figure}[!t]
  \centering
  \includegraphics[width=\linewidth]{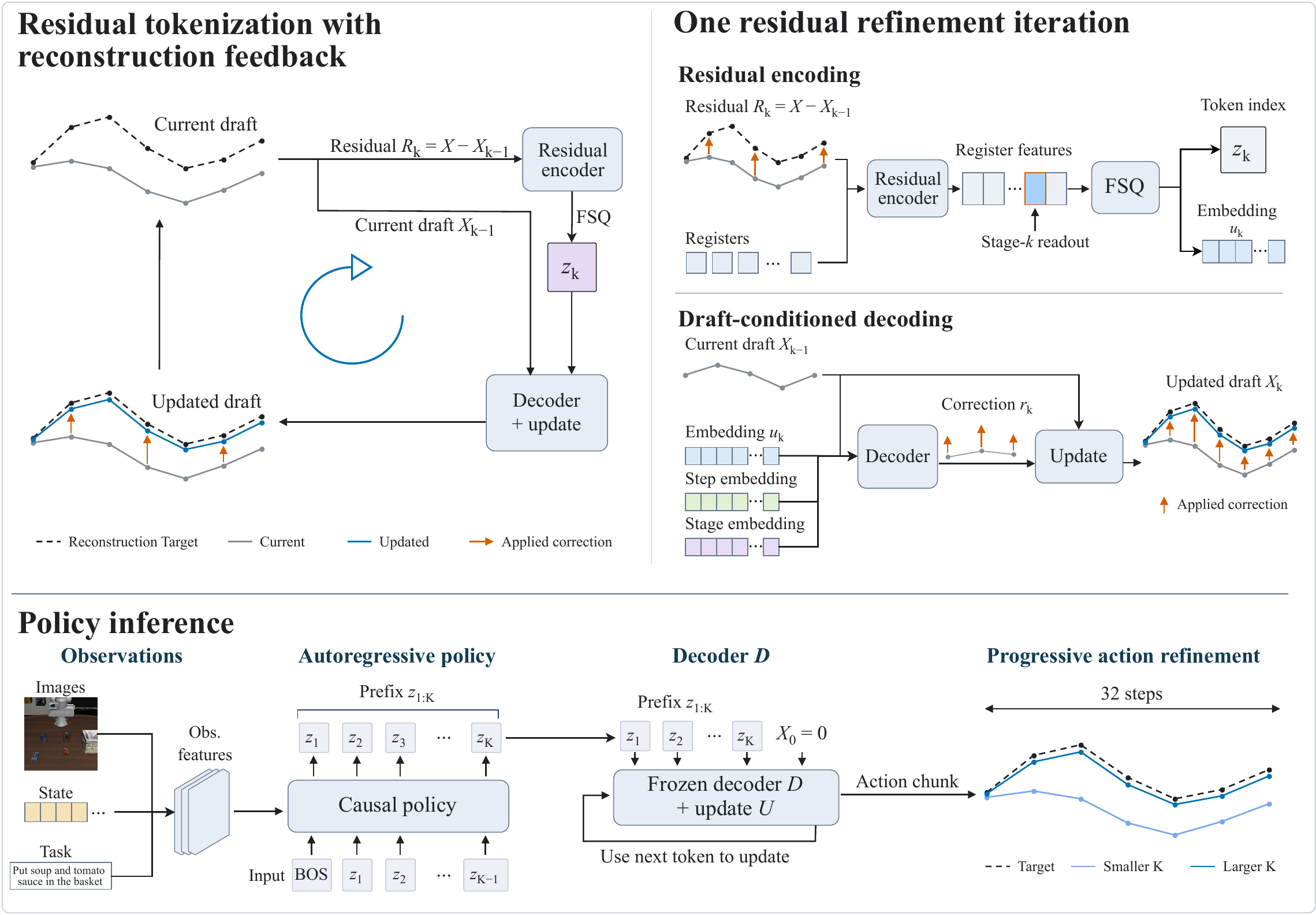}
  \caption{\textbf{Architecture of \method.} \textbf{Top left:} residual tokenization with reconstruction feedback across tokens. \textbf{Top right:} residual encoding and draft-conditioned decoding in one residual refinement iteration. \textbf{Bottom:} an observation-conditioned policy predicts tokens, whose prefixes are decoded by the frozen tokenizer.}
  \label{fig:codec}
\end{figure}

\section{Residual Action Tokenization}
\label{sec:method}
\method constructs action tokens through a reconstruction feedback loop (Figure~\ref{fig:codec}). Starting from a zero draft, it compares the current reconstruction with the expert action chunk, encodes the remaining error into one token, and uses that token to refine the draft. The updated reconstruction then determines what the next token must encode. The same encoder and decoder are reused at every stage. This loop constructs training labels; at deployment, a policy predicts the tokens from observations and only the decoder is needed to reconstruct actions.

\subsection{Residual Token Construction}
\label{sec:encoder}
\label{sec:coordinates}
The next token should describe what the decoded prefix has not yet reconstructed. Let $X\in\R^{H\times d}$ denote the expert chunk in reconstruction coordinates, and let $X_{k-1}$ be the draft produced by the first $k-1$ tokens in those same coordinates. With $X_0=0$, stage $k$ constructs a token from the remaining error:
\begin{equation}
  R_k=X-X_{k-1},\qquad
  (u_k,z_k)=\operatorname{FSQ}\!\left(E_\phi(R_k)_k\right).
  \label{eq:encoding}
\end{equation}
The Transformer encoder $E_\phi$ summarizes the residual using $M$ register features, one for each refinement stage. The subscript $k$ selects a stage-specific readout, allowing early and late stages to encode the residual differently. Finite scalar quantization (FSQ)~\citep{mentzer2024fsq} converts this feature into a discrete index $z_k$ and its quantized embedding $u_k$. The index is the policy's training label, while the embedding is the decoder's input. Only one token is produced per pass, not $M$ tokens at once. After decoding it, we rerun the encoder on the updated residual to construct the next token.

The encoder receives only the residual, not the draft as an additional input. Crucially, this residual is measured against the \emph{actual decoded reconstruction}: it includes errors left by quantization and decoding, rather than a discrepancy between latent features. Reconstruction feedback therefore gives each successive token a corrective role relative to the decoded prefix.

We normalize action channels using fixed training-set statistics and cumulatively sum translation commands within each chunk to expose accumulated command error. Rotation channels retain per-step values, while gripper controls are updated separately. These coordinates represent commands, not physical poses; cumulative sums start at zero and are inverted before execution. Appendix~\ref{app:coordinates} defines the transform and recurrent state.

\subsection{Reconstruction-Conditioned Refinement}
\label{sec:decoder}
The decoder interprets each new token in the context of what has already been reconstructed. It receives the token embedding $u_k$, the current draft $X_{k-1}$, and action-step and refinement-stage position features. A Transformer processes all $H$ action positions and predicts a correction proposal for the entire chunk. Consequently, the same token can produce different corrections for different drafts; it does not denote a fixed standalone motion.

We scale the $\tanh$-bounded proposal to limit each stage's correction magnitude, then update the recurrent draft while bounding decoded per-step commands to the normalized action range. Control channels accumulate corrections before bounding. These bounds make the applied change differ from the raw proposal. Appendix~\ref{app:decoder} gives the exact updates, and Table~\ref{tab:implementation} lists the proposal scales.

Every stage produces a complete $H$-step chunk. Additional tokens can revise any action step before execution, so token count controls refinement depth rather than chunk duration. With deterministic decoding and a fixed initial state, a prefix always gives the same reconstruction regardless of whether more tokens will follow. Retaining the recurrent state allows decoding to continue from that prefix without repeating earlier updates.

Algorithm~\ref{alg:tokenization} summarizes token construction. $D_\theta$ predicts the draft-conditioned proposal, and $\operatorname{Update}$ applies the bounded recurrent updates in Appendix~\ref{app:decoder}, returning the new draft and action-space reconstruction.

\begin{algorithm}[!htb]
  \caption{ReAT tokenization with reconstruction feedback}
  \label{alg:tokenization}
  \begin{algorithmic}[1]
    \Require Target chunk $X$ in reconstruction coordinates; token budget $M$
    \Ensure Tokens $z_{1:M}$ and prefix reconstructions $\widehat A^{(1)},\ldots,\widehat A^{(M)}$
    \State $X_0 \gets 0$; initialize decoder $\mathsf{state}$ to zero
    \For{$k=1,\ldots,M$}
      \State $R_k \gets X-X_{k-1}$
      \State $(u_k,z_k) \gets \operatorname{FSQ}(E_\phi(R_k)_k)$
      \State $r_k \gets D_\theta(u_k,X_{k-1},k)$
      \State $(X_k,\widehat A^{(k)},\mathsf{state}) \gets \operatorname{Update}(\mathsf{state},r_k)$
    \EndFor
    \State \Return $z_{1:M}$ and $\widehat A^{(1)},\ldots,\widehat A^{(M)}$
  \end{algorithmic}
\end{algorithm}

\subsection{Training for Progressive Reconstruction}
\label{sec:objective}
Residual feedback specifies what a token encodes, but does not by itself ensure that short prefixes reconstruct useful actions. We therefore train on reconstructions at different token budgets. We sample a prefix length $J$ uniformly from $1$ to $M$ and compare its decoded chunk with the expert actions. The reconstruction loss $\ell_{\mathrm{rec}}$ is a weighted combination of motion and control MSEs, measured after normalization in \emph{per-step action coordinates}, not cumulative coordinates. The objective is
\begin{equation}
  \mathcal L_{\mathrm{tok}}=
    \E_J\!\left[\ell_{\mathrm{rec}}(\widehat A^{(J)},A)\right]
    +\lambda_p\mathcal L_{\mathrm{progress}}+\mathcal L_{\mathrm{reg}}.
  \label{eq:loss}
\end{equation}
Uniform prefix sampling gives early and late reconstructions equal expected weight, rather than rewarding only the final chunk. The progress term penalizes increases in motion and control MSE between successive prefixes, encouraging additional tokens to improve reconstruction without guaranteeing monotonic improvement. The auxiliary regularizer $\mathcal L_{\mathrm{reg}}$ combines penalties on correction magnitude and temporal spread; it does not assign tokens to fixed action steps. Appendix~\ref{app:objective} expands these terms and Table~\ref{tab:implementation} lists their weights.

We jointly train the encoder and decoder through recurrent updates using straight-through gradients through FSQ. The decoded drafts provide both residual feedback and prefix supervision.

\subsection{Autoregressive Policy Learning and Inference}
\label{sec:policy}
After tokenizer training, we freeze the encoder, quantizer, decoder, and normalization statistics. Running the reconstruction feedback loop on each expert chunk produces deterministic target tokens $z^*_{1:M}$. An observation-conditioned policy then learns these tokens by teacher forcing:
\begin{equation}
  \mathcal L_{\mathrm{policy}}
    =-\E_{(O,c,A)}\sum_{k=1}^{M}\log\pi_\psi(z_k^*\mid O,c,z_{<k}^*).
  \label{eq:policy}
\end{equation}
The policy Transformer predicts all $M$ target tokens, conditioned on camera and proprioceptive features, task identity $c$, and preceding tokens.

At inference, the policy generates an externally selected budget of $K$ tokens, which the decoder converts into a complete action chunk without expert actions or residual encoding. The controller executes the first $H_e$ actions before replanning. Changing $K$ requires no policy retraining.

Token construction requires $M$ encoder and decoder passes, whereas deployment uses $K$ decoder passes in addition to policy generation. Section~\ref{sec:latency} reports the measured inference cost; Appendix~\ref{app:implementation} gives the implementation and deployment settings.

\section{Experiments}
\label{sec:experiments}
We evaluate control performance across token budgets, reconstruction quality, inference cost, and the contributions of individual design choices.

\subsection{Benchmarks, Data, and Evaluation Protocol}
\label{sec:evaluation-protocol}
\begin{figure}[!htb]
  \centering
  \begin{minipage}[t]{.42\linewidth}
    \vspace{0pt}
    \centering
    \includegraphics[width=\linewidth]{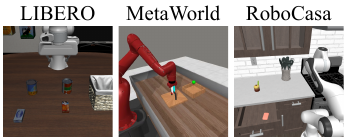}
    \caption{\textbf{Three benchmark domains.} Simulator renders (LIBERO, MetaWorld, RoboCasa).}
    \label{fig:benchmarks}
  \end{minipage}\hfill
  \begin{minipage}[t]{.55\linewidth}
    \vspace{0pt}
    \captionof{table}{\textbf{Data and evaluation scope.} $d$: action dimension. Episodes: total evaluation rollouts.}
    \label{tab:data}
    \centering\small
    \setlength{\tabcolsep}{2pt}
    \begin{tabular*}{\linewidth}{@{\extracolsep{\fill}}lrrrr@{}}
      \toprule
      Benchmark & Tasks & \shortstack{Demos/\\task} & $d$ & Episodes\\
      \midrule
      LIBERO-10 & 10 & 50 & 7 & 2500\\
      MetaWorld & 4 & 50 & 4 & 1000\\
      RoboCasa & 4 & \shortstack[r]{50 human +\\150 generated} & 12 & 1000\\
      \bottomrule
    \end{tabular*}
  \end{minipage}
\end{figure}

We evaluate \method on 19 manipulation tasks spanning LIBERO-10~\citep{liu2023libero}, MetaWorld~\citep{yu2020metaworld}, RoboCasa~\citep{nasiriany2024robocasa}, and real-world bucket stacking (Section~\ref{sec:real-world}). We train one multitask tokenizer and policy per simulation benchmark. Table~\ref{tab:data} summarizes the simulation data and evaluation scope, and Figure~\ref{fig:benchmarks} shows example environments. Appendix~\ref{app:protocol} details task identities, action interfaces, and dataset and simulator provenance.

\begin{table}[!t]
  \centering
  \begin{minipage}[t]{.49\linewidth}
    \vspace{0pt}
    \captionof{table}{\textbf{Policy success (\%).} Mean $\pm$ SEM across five evaluation rounds. Subscripts: token counts; bold: best mean. $*$: prefix length not sampled during OAT training.}
    \label{tab:main}
    \centering\small
    \setlength{\tabcolsep}{1pt}
    \renewcommand{\arraystretch}{1.12}
    \begin{tabular*}{\linewidth}{@{\extracolsep{\fill}}lccc@{}}
      \toprule
      Method & LIBERO & MetaWorld & RoboCasa\\
      \midrule
DP & $67.88\sem{0.83}$ & $27.90\sem{1.00}$ & $44.10\sem{0.80}$\\
Bin & $12.12\sem{0.29}$ & $19.70\sem{1.46}$ & $40.10\sem{1.95}$\\
FAST & $20.96\sem{0.64}$ & $20.80\sem{0.86}$ & $4.60\sem{0.99}$\\
QueST$_8$ & $65.52\sem{0.85}$ & $22.90\sem{0.53}$ & $52.70\sem{1.23}$\\
OAT$_2$ & $33.80\sem{0.81}$ & $27.00\sem{0.88}$ & $52.70\sem{1.21}$\\
OAT$_4$ & $50.96\sem{0.58}$ & $26.50\sem{1.92}$ & $53.30\sem{1.08}$\\
OAT$_6^*$ & $24.12\sem{0.39}$ & $26.70\sem{0.56}$ & $53.90\sem{1.85}$\\
OAT$_8$ & $61.48\sem{0.92}$ & $29.90\sem{1.39}$ & $53.60\sem{1.48}$\\
\midrule
\method{}$_{2}$ & $36.20\sem{1.28}$ & $19.00\sem{0.79}$ & $54.60\sem{2.00}$\\
\method{}$_{4}$ & $66.40\sem{0.75}$ & $27.10\sem{0.91}$ & $54.50\sem{2.63}$\\
\method{}$_{6}$ & $68.44\sem{1.38}$ & $28.80\sem{1.43}$ & $\mathbf{56.10}\sem{2.54}$\\
\method{}$_{8}$ & $\mathbf{71.16}\sem{0.68}$ & $\mathbf{31.70}\sem{1.02}$ & $55.90\sem{0.62}$\\
\bottomrule

    \end{tabular*}
  \end{minipage}\hfill
  \begin{minipage}[t]{.49\linewidth}
    \vspace{0pt}
      \captionof{table}{\textbf{Token count and end-to-end inference latency (ms).}}
  \label{tab:token-latency}
  \centering\small
  \setlength{\tabcolsep}{1pt}
  \renewcommand{\arraystretch}{1.12}
  \begin{tabular*}{\linewidth}{@{\extracolsep{\fill}}l*{3}{rr}@{}}
    \toprule
    & \multicolumn{2}{c}{LIBERO} & \multicolumn{2}{c}{MetaWorld} & \multicolumn{2}{c}{RoboCasa}\\
    \cmidrule(lr){2-3}\cmidrule(lr){4-5}\cmidrule(l){6-7}
    Policy & \#Tok. & Lat. & \#Tok. & Lat. & \#Tok. & Lat.\\
    \midrule
    DP & -- & 24.3 & -- & 23.9 & -- & 27.1\\
    Bin & 224 & 380.7 & 128 & 225.1 & 384 & 672.3\\
    FAST & 50.6 & 97.8 & 32.8 & 64.1 & 71.8 & 139.1\\
    QueST & 8 & 19.5 & 8 & 18.2 & 8 & 21.4\\
    OAT$_2$ & 2 & 9.5 & 2 & 7.8 & 2 & 12.1\\
    OAT$_4$ & 4 & 13.1 & 4 & 11.2 & 4 & 16.6\\
    OAT$_6$ & 6 & 16.6 & 6 & 14.7 & 6 & 20.0\\
    OAT$_8$ & 8 & 21.1 & 8 & 18.2 & 8 & 22.6\\
    \midrule
    \textbf{\method}$_2$ & 2 & 9.3 & 2 & 7.6 & 2 & 11.0\\
    \textbf{\method}$_4$ & 4 & 14.1 & 4 & 12.4 & 4 & 15.8\\
    \textbf{\method}$_6$ & 6 & 18.9 & 6 & 17.3 & 6 & 20.6\\
    \textbf{\method}$_8$ & 8 & 23.7 & 8 & 22.2 & 8 & 25.4\\
    \bottomrule
  \end{tabular*}

  \end{minipage}
\end{table}

All tokenizers and policies share a task-wise episode split, with 10\% held out for reconstruction evaluation and normalization statistics computed from training episodes only. All methods use observation horizon $H_o=2$, prediction horizon $H=32$, and execution horizon $H_e=16$. The main \method configuration uses a maximum of $M=8$ content tokens and a vocabulary of $V=1000$ codes. Main runs use batch size 256 and AdamW; Appendix~\ref{app:implementation} gives network dimensions, learning rates, and other implementation settings.

We conduct five complete simulation evaluation rounds for each trained policy, each comprising 50 episodes per task. Autoregressive decoding uses top-$k=10$ and temperature 1; Diffusion Policy uses 10 DDIM steps. Success is averaged equally over tasks within each round, and tables report the mean $\pm$ standard error of the mean (SEM) across the five round-level averages. Appendix~\ref{app:statistics} gives the estimator.

\subsection{Policy Performance and Token-Budget Trade-offs}
\label{sec:success}
\label{sec:diagnostic}
We first test whether \method supports effective control with compact action sequences and how task success changes when the token budget varies. We compare against OAT, QueST, FAST, scalar binning (Bin), and Diffusion Policy (DP), with success rates reported in Table~\ref{tab:main}. OAT and \method are evaluated at two, four, six, and eight tokens; QueST uses eight tokens, while FAST and Bin retain their native sequence lengths. OAT's nested dropout trains power-of-two prefixes, so its six-token result tests an intermediate budget not explicitly supervised during training.

At eight tokens, \method outperforms the evaluated baselines on all three benchmarks, exceeding the strongest baseline by 3.28, 1.80, and 2.00 percentage points on LIBERO, MetaWorld, and RoboCasa, respectively.

\begin{figure}[!t]
  \centering
  \begin{minipage}[t]{.52\linewidth}
    \vspace{0pt}
    \centering
    \includegraphics[width=.827\linewidth]{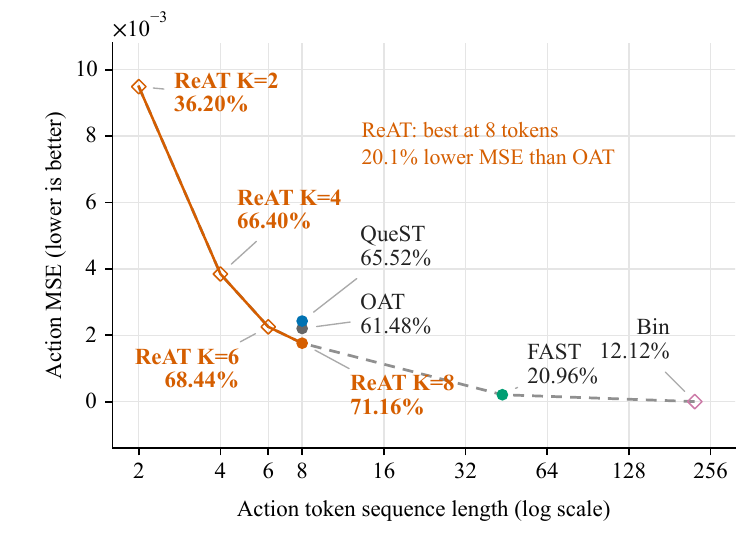}
    \caption{\textbf{LIBERO reconstruction error versus token length.} Raw-action MSE on validation chunks; percentages give policy success. Dashed links: observed Pareto front. FAST uses mean token length.}
    \label{fig:libero-pareto}
  \end{minipage}\hfill
  \begin{minipage}[t]{.46\linewidth}
    \vspace{0pt}
    \captionof{table}{\textbf{Real-world bucket stacking.} Twenty trials per configuration.}
    \label{tab:real-world}
    \centering\small
    \setlength{\tabcolsep}{3pt}
    \begin{tabular*}{\linewidth}{@{\extracolsep{\fill}}lrr@{}}
      \toprule
      Method & \shortstack{Successes /\\trials} & Success (\%)\\
      \midrule
      DP & 0/20 & 0\\
      Bin & 0/20 & 0\\
      FAST & 0/20 & 0\\
      QueST$_8$ & 7/20 & 35\\
      OAT$_8$ & 8/20 & 40\\
      \midrule
      \method{}$_2$ & 6/20 & 30\\
      \method{}$_4$ & 5/20 & 25\\
      \method{}$_6$ & 12/20 & 60\\
      \method{}$_8$ & 14/20 & \textbf{70}\\
      \bottomrule
    \end{tabular*}
  \end{minipage}
\end{figure}

Varying \method's budget over $K\in\{2,4,6,8\}$ without retraining increases success on LIBERO and MetaWorld, while RoboCasa varies little; its six-token mean exceeds the eight-token mean by only 0.20 percentage points, small relative to the SEMs. Prefix supervision optimizes reconstruction, so additional tokens need not improve task success monotonically. At the shared trained budgets $K\in\{2,4,8\}$, \method outperforms OAT on LIBERO and RoboCasa, with the largest LIBERO gap at four tokens (66.40\% versus 50.96\%). MetaWorld OAT performs better at two tokens (27.00\% versus 19.00\%), so the full-budget advantage does not extend to every short-prefix setting.

\paragraph{Deployment cost.}
\label{sec:latency}
Fewer tokens reduce autoregressive generation steps, but \method also incurs recurrent decoding cost. We measure end-to-end inference latency for all methods under the same protocol on a single NVIDIA RTX 6000D at batch size one, including observation encoding, token generation, and action decoding. Table~\ref{tab:token-latency} reports the results. Eight-token \method takes 22.2--25.4\,ms versus OAT's 18.2--22.6\,ms; reducing \method to two tokens lowers latency to 7.6--11.0\,ms without shortening the action horizon. Together with Table~\ref{tab:main}, these results characterize the control--latency trade-off rather than a uniform speed advantage. Appendix~\ref{app:latency} details the timing protocol.

\paragraph{Reconstruction quality.}
To determine whether additional tokens improve reconstruction independently of policy prediction, we encode held-out demonstration actions and decode their prefixes. Figure~\ref{fig:libero-pareto} compares LIBERO reconstruction MSE against token length. MSE is measured in the original action coordinates rather than the cumulative representation used internally, averaging first over chunks within each task and then equally across tasks. Appendix~\ref{app:dimension-reconstruction} provides per-dimension $R^2$ and token counts.

Increasing \method's budget from two to eight tokens reduces MSE from $0.009491$ to $0.001763$, reaching $20.1\%$ below OAT at eight tokens. FAST and Bin obtain lower MSE with longer sequences but lower policy success (Table~\ref{tab:main}), showing that reconstruction fidelity alone does not determine control performance.

\subsection{Real-World Bucket Stacking}
\label{sec:real-world}
\begin{figure}[!htb]
  \centering
  \begin{minipage}[t]{.49\linewidth}
    \centering
    {\small\textbf{ReAT: successful sequence}}\par\vspace{3pt}
    \begin{minipage}[t]{.32\linewidth}
      \centering
      \includegraphics[width=\linewidth]{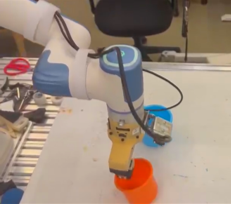}\\[-2pt]
      {\footnotesize Approach}
    \end{minipage}\hfill
    \begin{minipage}[t]{.32\linewidth}
      \centering
      \includegraphics[width=\linewidth]{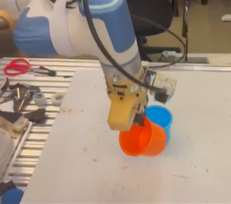}\\[-2pt]
      {\footnotesize Lift}
    \end{minipage}\hfill
    \begin{minipage}[t]{.32\linewidth}
      \centering
      \includegraphics[width=\linewidth]{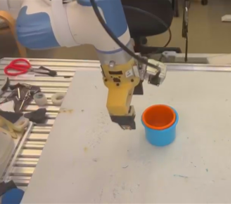}\\[-2pt]
      {\footnotesize Stack and release}
    \end{minipage}
  \end{minipage}\hfill
  \begin{minipage}[t]{.49\linewidth}
    \centering
    {\small\textbf{Baseline failure examples}}\par\vspace{3pt}
    \begin{minipage}[t]{.32\linewidth}
      \centering
      \includegraphics[width=\linewidth]{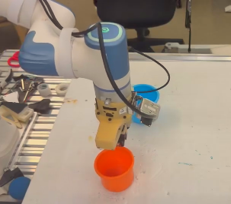}\\[-2pt]
      {\footnotesize DP}
    \end{minipage}\hfill
    \begin{minipage}[t]{.32\linewidth}
      \centering
      \includegraphics[width=\linewidth]{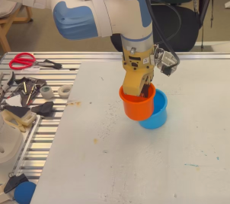}\\[-2pt]
      {\footnotesize OAT}
    \end{minipage}\hfill
    \begin{minipage}[t]{.32\linewidth}
      \centering
      \includegraphics[width=\linewidth]{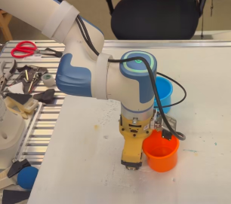}\\[-2pt]
      {\footnotesize QueST}
    \end{minipage}
  \end{minipage}
  \caption{\textbf{Real-world bucket stacking.} Left: a successful \method sequence on the X-Trainer arm. Right: failure examples in which DP closes its gripper prematurely and descends into the bucket without grasping it, OAT grasps the bucket but misaligns it during placement, and QueST fails to align the gripper for a successful grasp.}
  \label{fig:real-world-stack}
\end{figure}

To test whether the representation supports physical manipulation, we evaluate bucket stacking on an X-Trainer robotic arm (Figure~\ref{fig:real-world-stack}). Policies are trained on 90 demonstration episodes using a head-mounted camera and one wrist-mounted camera. Bucket positions are randomized before each autonomous trial. Success requires the buckets to remain stacked after gripper release within 30 seconds, without human intervention. We evaluate \method at two, four, six, and eight tokens, and OAT and QueST at eight tokens; Table~\ref{tab:real-world} reports 20 trials per configuration.

Eight-token \method achieves 70\% success, compared with 40\% for OAT and 35\% for QueST. Larger budgets generally help, although success is not monotonic across budgets.

Additional observations show that Bin repeatedly approaches the bucket rim without aligning the grasp. FAST encounters decoding failures when policy-generated token sequences yield an incorrect number of action coefficients. A held-out validation check confirms this failure mode in policy generation, while tokens encoded from expert actions decode successfully (Appendix~\ref{app:fast-validity}).

\begin{table}[!t]
\centering
\begin{minipage}[t]{.42\linewidth}
  \vspace{0pt}
  \caption{\textbf{FSQ configuration sensitivity.} LIBERO success (\%) $\pm$ SEM at $K=8$.}
  \label{tab:capacity}
  \centering\small
  \setlength{\tabcolsep}{2pt}
  \begin{tabular*}{\linewidth}{@{\extracolsep{\fill}}lrr@{}}
    \toprule
    FSQ levels & Codes & Success\\
    \midrule
$[8,6,5]$ & 240 & $61.76\sem{0.55}$\\
$[8,8,8]$ & 512 & $65.16\sem{0.95}$\\
$[8,5,5,5]$ & 1000 & $\mathbf{71.16}\sem{0.68}$\\
$[8,8,6,5]$ & 1920 & $68.28\sem{0.88}$\\
$[7,5,5,5,5]$ & 4375 & $66.80\sem{0.46}$\\
\bottomrule

  \end{tabular*}
\end{minipage}\hfill
\begin{minipage}[t]{.55\linewidth}
  \vspace{0pt}
  \captionof{table}{\textbf{LIBERO mechanism ablations.} Success (\%) $\pm$ SEM; each variant uses the same policy across token budgets. Bold: best mean per budget.}
  \label{tab:mechanism-success}
  \centering\small
  \setlength{\tabcolsep}{1.5pt}
  \renewcommand{\arraystretch}{1.18}
  \begin{tabular*}{\linewidth}{@{\extracolsep{\fill}}ccccc@{}}
    \toprule
    $K$ & Full & \shortstack{Feedback-\\off} & \shortstack{Full-length\\only} & \shortstack{Direct\\actions}\\
    \midrule
2 & $\mathbf{36.20}\sem{1.28}$ & $25.32\sem{0.48}$ & $0.04\sem{0.04}$ & $33.68\sem{1.42}$\\
4 & $\mathbf{66.40}\sem{0.75}$ & $51.32\sem{0.83}$ & $1.76\sem{0.33}$ & $55.88\sem{0.50}$\\
6 & $\mathbf{68.44}\sem{1.38}$ & $58.72\sem{0.73}$ & $8.52\sem{0.69}$ & $55.44\sem{1.09}$\\
8 & $\mathbf{71.16}\sem{0.68}$ & $60.24\sem{1.02}$ & $70.80\sem{0.68}$ & $56.84\sem{1.05}$\\
\bottomrule

  \end{tabular*}
\end{minipage}
\end{table}

\subsection{Ablation Studies}

\paragraph{Mechanism ablations.}
The main comparison evaluates the complete design; to examine the roles of feedback, action coordinates, and prefix supervision, we ablate these components on LIBERO. Table~\ref{tab:mechanism-success} reports success under the main training protocol and batch size. Each tokenizer and policy is trained from scratch and evaluated across prefix budgets.

Feedback-off tests the contribution of reconstruction feedback by encoding the target $X$ once and quantizing all $M$ register features, $(u_{1:M},z_{1:M})=\mathrm{FSQ}(E_\phi(X))$, without recomputing encoder inputs from intermediate drafts. It retains the encoder architecture, quantizer, draft-conditioned recurrent decoder, and tokenizer loss. Direct actions tests the cumulative representation by setting $\mathcal I=\varnothing$ while retaining feedback; rotation and gripper channels remain direct. Full-length only tests the need for intermediate-prefix supervision by setting random-prefix and progress-loss weights to zero and the full-length reconstruction weight to one.

Feedback-off reduces eight-token success from 71.16\% to 60.24\% and performs worse at every tested budget. Direct actions reaches 56.84\%, supporting a separate contribution from cumulative translation; Appendix~\ref{app:matched-coordinates} provides a coordinate-matched OAT comparison. Full-length only retains similar eight-token performance but performs poorly at shorter prefixes, supporting the need for prefix supervision.

\paragraph{FSQ configuration and capacity.}
To test quantization capacity at a fixed eight-token budget, we compare five FSQ configurations on LIBERO (Table~\ref{tab:capacity}). The 1000-code configuration performs best; increasing capacity further does not improve policy success.

\section{Discussion and Limitations}
Reconstruction feedback gives tokens a corrective role, not a fixed motion meaning: the same symbol can produce different updates for different drafts. The ablations support both feedback and prefix supervision, while showing that reconstruction fidelity alone does not determine control performance.

Budgets are externally selected and chunks are generated before execution; learned stopping, asynchronous refinement, and inter-chunk continuity remain open directions. Output bounds do not guarantee safe motion, nor does feedback eliminate policy prediction errors. Physical evaluation covers one task with 20 trials per configuration; additional tasks and hardware platforms would clarify deployment generality.

\section{Conclusion}
\method constructs action tokens from the residual left by the decoded prefix and converts them into draft-conditioned corrections. Prefix supervision makes this representation usable at multiple token budgets with the same autoregressive policy and action horizon. Experiments across three simulation benchmarks and real-world bucket stacking demonstrate its control performance, while LIBERO ablations support reconstruction feedback and prefix supervision. More broadly, these findings highlight explicit residual correction as a design principle for action tokenization, extending the goal beyond compact encoding to structuring how successive tokens contribute to action reconstruction.

\clearpage
\section*{Ethics Statement}
This work uses existing robot demonstration datasets, simulation, and a physical bucket-stacking task, with no new human-subject study. Physical deployment requires robot-specific safety measures beyond bounded action outputs. Dataset reuse is subject to the original licenses and attribution requirements.

\section*{Reproducibility Statement}
Sections~\ref{sec:method} and~\ref{sec:experiments} specify the method, objectives, and evaluation protocol. Algorithm~\ref{alg:tokenization} summarizes token construction; Appendix~\ref{app:implementation} provides architecture and optimization settings, exact update equations, and label-construction details. Appendices~\ref{app:protocol} and~\ref{app:statistics} describe the data interfaces and uncertainty estimator; Appendix~\ref{app:latency} records the timing protocol and model provenance.

\section*{Statement on the Use of AI}
Generative AI tools, including OpenAI Codex, were used to polish writing, retrieve and discover related literature, assist with research ideation and execution (experiment planning and monitoring), and draft sections of the manuscript.

\section*{Acknowledgments}
This work was supported by Guangdong Provincial Natural Science Foundation for Outstanding Youth Team Project (No.~2024B1515040010), National Natural Science Foundation of China under Grants U23A20391 and 62372188, Guangdong Natural Science Foundation under Grant 2024A1515010100, and Pivotal Technology Research Project of the Hengqin Guangdong-Macao In-Depth Cooperation Zone under Grant 2530004025492.

\bibliography{references}
\bibliographystyle{iclr2027_conference}
\clearpage
\appendix
\section{Implementation Details}
\label{app:implementation}
\subsection{Architecture and Optimization}
Table~\ref{tab:implementation} records the main \method configuration.

\begin{table}[htbp]
  \caption{\textbf{Main \method settings.} EMA is enabled in both stages.}
  \label{tab:implementation}
  \centering\small
  \begin{tabular}{ll}
    \toprule
    Setting & Value\\
    \midrule
    Observation / prediction / execution horizon & 2 / 32 / 16\\
    Maximum content tokens / vocabulary & 8 / 1000\\
    FSQ scalar levels & $[8,5,5,5]$\\
    Register encoder & width 256, head dimension 64\\
    Edit decoder & width 256, head dimension 64\\
    Autoregressive policy & 4 layers, width 256, 4 heads\\
    Encoder / decoder / policy dropout & 0.1 / 0.1 / 0.1\\
    Proposal scales $\alpha$ / $\alpha_g$ & 8 / 4\\
    Full / random-prefix reconstruction weight & 0 / 1; uniform lengths 1--8\\
    Progress / control MSE weight $\lambda_p$ / $\lambda_g$ & 0.2 / 0.1\\
    Locality / edit weight $\lambda_{\mathrm{loc}}$ / $\lambda_{\mathrm{edit}}$ & $10^{-3}$ / $10^{-4}$\\
    Batch size, each training stage & 256\\
    Optimizer & AdamW, $\beta=(0.9,0.95)$, no weight decay\\
    Tokenizer / policy / observation learning rate & $5\!\times\!10^{-5}$ / $5\!\times\!10^{-5}$ / $10^{-5}$\\
    Schedule / warmup / gradient norm cap & Constant / 100 steps / 1.0\\
    Frozen teacher label precision / mode & FP32 / evaluation mode\\
    Policy and visual encoder training & BF16 mixed precision\\
    \bottomrule
  \end{tabular}
\end{table}

Tokenizer dropout is active during training. The intermediate reconstructions used to define encoder residuals also supply the prefix losses, and gradients flow through the recurrent updates and straight-through quantizer.

\paragraph{Visual encoder.}
Each camera uses a separate ResNet-18 convolutional backbone initialized without pretrained weights. Batch normalization is replaced by GroupNorm, and spatial softmax pooling with 32 keypoints produces a 64-dimensional feature per camera. The visual encoder is optimized jointly with the policy at learning rate $10^{-5}$; it is not frozen. Appendix~\ref{app:observations} describes image cropping and evaluation preprocessing.

\paragraph{Baseline configurations.}
The simulation policies use four-layer, width-256 action Transformers with four attention heads. We use the tokenizer architectures described in the OAT and QueST papers~\citep{liu2026oat,mete2024quest}, trained under the protocol described in Appendix~\ref{app:training}. FAST fits a benchmark-specific byte-pair vocabulary of 1024 entries with DCT quantization scale 10. Bin uses scalar binning configured with 256 bins over normalized actions. DP uses a cosine noise schedule with 100 training timesteps and predicts noise, with 10 DDIM steps at inference.

\subsection{Training Duration and Checkpoint Evaluation}
\label{app:training}
Tokenizer training stops at a stage boundary when the minimum validation loss decreases by less than 0.5\% between two consecutive 100-epoch windows. \method, OAT and QueST use validation reconstruction MSE. The checkpoint minimizing the corresponding criterion is frozen before policy training. FAST fits its vocabulary without gradient-based tokenizer training, Bin uses fixed scalar bins, and DP has no tokenizer-training stage. Policies are trained for 500 epochs, and checkpoints are selected by the lowest validation loss: cross-entropy for autoregressive policies and denoising MSE for DP. Evaluation follows Section~\ref{sec:evaluation-protocol}.

\subsection{Action Coordinates and Recurrent State}
\label{app:coordinates}
Let $\widetilde A=N(A)$ be the expert chunk after per-channel affine normalization. For each action channel $j$, training-episode extrema $a_j^{\min}$ and $a_j^{\max}$ define
\[
  N(A)_{t,j}=2\frac{A_{t,j}-a_j^{\min}}{a_j^{\max}-a_j^{\min}}-1.
\]
The same statistics are retained for validation and deployment. Channels with range below $10^{-4}$ use unit scale and offset $-a_j^{\min}$ to avoid division by a near-zero range. The inverse affine map restores the benchmark action units after decoding.

We partition normalized action dimensions into motion channels $\mathcal C$ and separately updated control channels $\mathcal G$, including gripper commands. Translation channels $\mathcal I\subseteq\mathcal C$ use cumulative sums; rotation channels retain per-step values. The transform $C_{\mathcal I}$ and its inverse $B_{\mathcal I}$ are
\begin{equation}
  (C_{\mathcal I}x)_{t,j}=\begin{cases}\sum_{s=1}^{t}x_{s,j},&j\in\mathcal I,\\x_{t,j},&j\notin\mathcal I,\end{cases}
  \quad
  (B_{\mathcal I}q)_{t,j}=\begin{cases}q_{t,j}-q_{t-1,j},&j\in\mathcal I,\\q_{t,j},&j\notin\mathcal I,\end{cases}
  \label{eq:coordinates}
\end{equation}
with $q_{0,j}=0$, where $t$ indexes action steps and $j$ indexes channels. The motion target is $Q=C_{\mathcal I}\widetilde A_{\mathcal C}$, and the control target is $g=\widetilde A_{\mathcal G}$. Cumulative sums restart at zero per chunk and represent normalized commands, not physical poses. Appendix~\ref{app:protocol} specifies the channel groups.

The recurrent state comprises a motion draft $S^{(k)}$ and control preactivations $L^{(k)}$. After $k$ tokens, $S^{(k)}$ reconstructs $Q$, while $\tanh L^{(k)}$ reconstructs $g$. Control corrections accumulate in $L^{(k)}$ before $\tanh$ bounds their output. Both states start at zero. The target and visible draft used in Equation~\ref{eq:encoding} are
\begin{equation}
  X=\operatorname{Concat}(Q,g),\qquad
  X_k=\operatorname{Concat}\!\left(S^{(k)},\tanh L^{(k)}\right).
\end{equation}
Here $\operatorname{Concat}$ joins channels in their original order, not along time; $X,X_k\in\R^{H\times d}$ and $X_0=0$. The decoder retains $L^{(k)}$ in its recurrent state even though its visible draft contains the bounded value $\tanh L^{(k)}$.

\subsection{Encoder Features and Bounded Decoder Updates}
\label{app:decoder}
The Transformer encoder $E_\phi$ produces $M$ register feature vectors from each residual $R_k$. Only vector $k$ is quantized at stage $k$, yielding the index $z_k$ and embedding $u_k$ in Equation~\ref{eq:encoding}. The encoder is rerun on the updated residual at each stage.

For action step $t$ and token stage $k$, the decoder combines the draft row $X_{k-1,t}$ with $u_k$ and positional features:
\begin{equation}
  h_{k,t}=p_t^{\mathrm{time}}+W_s X_{k-1,t}
                   +W_u u_k+p_k^{\mathrm{tok}}+s_k,\qquad
  r_k=\tanh\!\left(W_r F_\theta(h_k)\right).
  \label{eq:features}
\end{equation}
Here $p_t^{\mathrm{time}}$ and $p_k^{\mathrm{tok}}$ encode action and token positions, $s_k$ is a learned stage embedding, and $h_k$ collects all $H$ input vectors. The Transformer $F_\theta$ operates across action positions; the $W$ matrices are learned projections. Its bounded proposal $r_k\in\R^{H\times d}$ covers all action steps.

For motion channels, we add the proposal, convert the temporary draft $\overline S^{(k)}$ to per-step commands, bound them with $\tanh$, and transform back. Control channels accumulate corrections before $\tanh$. The normalized predictions $\widehat a^{(k)}$ are
\begin{align}
  \overline S^{(k)} &= S^{(k-1)}+\alpha r_{k,\mathcal C}, &
  \widehat a^{(k)}_{\mathcal C} &= \tanh\!\left(B_{\mathcal I}\overline S^{(k)}\right), &
  S^{(k)} &= C_{\mathcal I}\widehat a^{(k)}_{\mathcal C}, \label{eq:continuous}\\
  L^{(k)} &= L^{(k-1)}+\alpha_g r_{k,\mathcal G}, &
  \widehat a^{(k)}_{\mathcal G} &= \tanh L^{(k)}, && \label{eq:discrete}
\end{align}
where $\alpha$ and $\alpha_g$ scale the proposals. Undoing normalization gives $\widehat A^{(k)}=N^{-1}(\widehat a^{(k)})$. Deployment also applies empirical action bounds and benchmark-controller clipping. Because of bounding, the actual draft change $X_k-X_{k-1}$ need not equal the scaled proposal. Retaining $(S^{(K)},L^{(K)})$ permits further decoding without repeating the first $K$ updates.

\subsection{Expanded Tokenizer Objective}
\label{app:objective}
The motion and control reconstruction errors at prefix $k$ are
\begin{equation}
  e_k=\frac{\norm{\widehat a^{(k)}_{\mathcal C}-\widetilde A_{\mathcal C}}_F^2}
                    {H|\mathcal C|},\qquad
  g_k=\frac{\norm{\tanh L^{(k)}-\widetilde A_{\mathcal G}}_F^2}{H|\mathcal G|}.
\end{equation}
Both are MSEs in normalized per-step action coordinates; $\|\cdot\|_F$ denotes the Frobenius norm. The reconstruction term in Equation~\ref{eq:loss} is $\ell_{\mathrm{rec}}(\widehat A^{(k)},A)=e_k+\lambda_g g_k$. For $J\sim\operatorname{Uniform}\{1,\ldots,M\}$, the expanded objective is
\begin{equation}
  \begin{aligned}
    \mathcal L_{\mathrm{tok}}={}&\E_J[e_J+\lambda_g g_J]
      +\frac{\lambda_p}{M-1}\sum_{k=1}^{M-1}\left([e_{k+1}-e_k]_++\lambda_g[g_{k+1}-g_k]_+\right)\\
      &+\lambda_{\mathrm{loc}}\mathcal L_{\mathrm{loc}}
      +\lambda_{\mathrm{edit}}\mathcal L_{\mathrm{edit}}.
  \end{aligned}
  \label{eq:loss-expanded}
\end{equation}
Training estimates the expectation by uniform prefix sampling. The progress term $\mathcal L_{\mathrm{progress}}$ is the average of the bracketed penalties, with $[x]_+=\max(x,0)$; increases are penalized separately for motion and control errors. The auxiliary term is $\mathcal L_{\mathrm{reg}}=\lambda_{\mathrm{loc}}\mathcal L_{\mathrm{loc}}+\lambda_{\mathrm{edit}}\mathcal L_{\mathrm{edit}}$. Table~\ref{tab:implementation} gives all weights.

\paragraph{Applied-correction regularization.}
Let $\delta_k$ concatenate $S^{(k)}-S^{(k-1)}$ and $\tanh L^{(k)}-\tanh L^{(k-1)}$. For $k\geq2$, define $w_{k,t}=\sqrt{\norm{\delta_{k,t}}_2^2+10^{-6}}$. The locality term averages $(\sum_t w_{k,t})^2/(H\sum_t w_{k,t}^2)$, and the edit term averages squared elements of $\delta_k$. These regularizers encourage moderate, concentrated corrections without fixing their temporal support. They operate after the bounding projection; because of this projection, a zero proposal is not generally an identity update.

\subsection{Token Construction and Deployment Details}
In Algorithm~\ref{alg:tokenization}, the input $X$ is the normalized, coordinate-transformed expert chunk defined in Appendix~\ref{app:coordinates}. The internal decoder state is $(S^{(k)},L^{(k)})$, and $D_\theta$ denotes the edit proposal in Equation~\ref{eq:features}. The $\operatorname{Update}$ operation applies Equations~\ref{eq:continuous}--\ref{eq:discrete}, forms $X_k$, and undoes normalization to produce $\widehat A^{(k)}$. Tokenizer training differentiates through this recurrent computation with straight-through quantization and the prefix objective in Equation~\ref{eq:loss}. Policy-label construction instead runs the frozen encoder, quantizer, and decoder in FP32 with dropout disabled.

At deployment, a beginning-of-sequence symbol initializes the policy, which generates $K$ content tokens while masking BOS and STOP from its output vocabulary. Each integer code maps to its FSQ embedding; the decoder applies the same proposal and update operations from a zero-initialized state, without expert actions or residual encoding. The reconstructed chunk is clamped to the empirical and benchmark-controller action bounds. The controller executes its first 16 actions, obtains new observations, and repeats inference.

\section{Benchmark and Evaluation Details}
\label{app:protocol}
\subsection{Tasks and Action Interfaces}
LIBERO uses the original LIBERO-10 task suite and its seven-dimensional action interface. MetaWorld uses Box Close, Coffee Pull, Disassemble, and Stick Pull. RoboCasa uses Close Drawer, Coffee Press Button, Turn Off Microwave, and Turn Off Sink Faucet.

The RoboCasa dataset combines 50 human demonstrations and 150 machine-generated demonstrations per task. We evaluate in RoboCasa v0.2 using its twelve-dimensional action interface, with dataset action channels reordered to the simulator controller's expected order.

The separately updated control indices $\mathcal G$ (zero-based) are $[6]$ for LIBERO, $[3]$ for MetaWorld, and $[4,11]$ for RoboCasa. The integrated translation indices $\mathcal I$ are $[0,1,2]$ for LIBERO and MetaWorld, and $[5,6,7]$ for RoboCasa in dataset order. Rotation and all remaining channels retain direct per-step action coordinates. All twelve RoboCasa dimensions are retained before the controller permutation.

The main method uses per-channel affine action normalization and the action representation defined in Appendix~\ref{app:coordinates}.

\subsection{Observation Processing and Reset Handling}
\label{app:observations}
All compared methods use the same benchmark-specific camera and proprioceptive inputs. Simulation images are randomly cropped to $76\times76$ during training and center-cropped to $76\times76$ during validation and rollout evaluation before ResNet encoding. For environment initialization, the MetaWorld evaluator clears residual bottle velocity after reset, while LIBERO lets the scene settle before acquiring the initial observation.

OAT's short-prefix evaluations use the same policy and tokenizer as its eight-token result within each benchmark, without retraining or prefix-specific model selection. LIBERO, MetaWorld and RoboCasa use center cropping and the same evaluation seed set as \method. Each added budget uses five rounds of 50 episodes per task; the eight-token results are retained from the existing evaluations.

\subsection{Statistical Reporting}
\label{app:statistics}
Let $Y_{t,r,i}\in\{0,1\}$ denote success for task $t$, evaluation round $r\in\{1,\ldots,5\}$, and episode $i\in\{1,\ldots,50\}$. We first average episodes and tasks within each round, then average the five round-level means:
\begin{equation}
 \widehat p_r=\frac{1}{T}\sum_{t=1}^{T}\left(\frac{1}{50}\sum_{i=1}^{50}Y_{t,r,i}\right),
 \qquad \widehat p=\frac{1}{5}\sum_{r=1}^{5}\widehat p_r.
\end{equation}
Tables~\ref{tab:main} and~\ref{tab:mechanism-success} report $\widehat p$ and the standard error across round-level means:
\begin{equation}
 s_{\mathrm{round}}=\sqrt{\frac{1}{4}\sum_{r=1}^{5}(\widehat p_r-\widehat p)^2},
 \qquad \mathrm{SEM}=\frac{s_{\mathrm{round}}}{\sqrt{5}}.
\end{equation}
Both $\widehat p$ and SEM are multiplied by 100 for percentage reporting.

\subsection{FAST Decoding Validity on Bucket Stacking}
\label{app:fast-validity}
We examine all 300 validation windows from nine held-out bucket-stacking demonstrations using the exported FAST policy and tokenizer. The check verifies that the exported policy weights match the saved EMA weights. For each window, we separately decode tokens obtained by encoding the expert action chunk and tokens freely generated by the observation-conditioned policy, without supplying expert token prefixes. This FAST configuration predicts 16 steps with seven action dimensions, requiring exactly 112 DCT coefficients after token decoding.

All 300 expert-action token sequences decode successfully. In contrast, policy generation with top-$k=10$ and temperature 1 produces 263/300 invalid sequences (87.67\%); greedy generation produces 251/300 (83.67\%). Every invalid sequence has an incorrect coefficient count and cannot be reshaped into the required action chunk. These results identify a failure of policy-generated sequence validity in this check, rather than a failure to decode tokens encoded from expert actions. The percentages describe offline validation chunks, separately from the physical trial success rates in Table~\ref{tab:real-world}.

\section{Inference-Latency Measurement Protocol}
\label{app:latency}
We measure synchronized host-wall time on one RTX 6000D with batch size one, gradients disabled, and FP32 parameters (PyTorch 2.10.0, CUDA 12.8). Each benchmark cycles through 20 GPU-resident observation windows. Each of five repetitions consists of 50 warmup calls followed by 100 timed calls. Table~\ref{tab:token-latency} reports the mean across all timed calls. Timing includes observation encoding, token generation, action decoding, and final synchronization; it excludes simulator execution, RPC, and external data loading. Jobs run serially with one CPU thread pinned to a single core and GPU isolation checked during measurement.

All methods are timed under eager execution. \method retains fixed-prefix decoding, cached static features, and final-prefix-only output computation: it executes only the requested $K$ stages without changing the decoded prefix.

Native action backbones have four layers, width 256, and four heads. DP uses 10 DDIM steps; Bin uses 256 bins per scalar; autoregressive baselines use KV caching, top-$k=10$, and temperature 1. The \method timing variants share parameters and differ only in the requested token budget. FAST timings replay valid demonstration codes through the autoregressive generation path and include CPU action decoding, so its token counts are demonstration-conditioned rather than policy-predicted stopping lengths.

OAT's short-prefix measurements generate exactly $K\in\{2,4,6\}$ tokens and decode a complete 32-step chunk, replacing the unused latent positions with the learned dropout mask. Each OAT budget uses the same five-repeat protocol and physical GPU as the other rows.

\section{Per-Dimension Tokenizer Reconstruction}
\label{app:dimension-reconstruction}
Figure~\ref{fig:libero-pareto} reports LIBERO reconstruction MSE, which depends on channel scale. To compare translation, rotation, and gripper quality, Table~\ref{tab:dimension-reconstruction} reports per-dimension $R^2$ on the same validation split, normalized by target variance within each task and dimension. For task $t$ and dimension $j$, we compute
\begin{equation}
  R^2_{t,j}=1-\frac{\sum_i(a_{t,i,j}-\widehat a_{t,i,j})^2}
                       {\sum_i(a_{t,i,j}-\bar a_{t,j})^2},
  \qquad \bar a_{t,j}=\frac{1}{n_t}\sum_i a_{t,i,j},
\end{equation}
Here $i$ indexes action steps in held-out chunks, $n_t$ is their count, and $\widehat a$ comes from expert-action tokens rather than policy predictions. $R^2=1$ denotes exact reconstruction; $0$ matches the task-specific channel mean; negative values are worse. We average equally across the ten tasks and seven dimensions. Every task--dimension pair has nonzero target variance.

\begin{table}[!htbp]
  \caption{\textbf{LIBERO tokenizer reconstruction by action dimension.} $R^2$ values are task-averaged; MSE is measured in original action coordinates. $\bar K$ counts content tokens.}
  \label{tab:dimension-reconstruction}
  \centering\small
  \begin{tabular*}{\linewidth}{@{\extracolsep{\fill}}lrrrrr@{}}
    \toprule
    Action dimension / summary & \method & OAT & QueST & FAST & Bin\\
    \midrule
    Translation $x$ & 0.94889 & 0.93819 & 0.94711 & 0.99463 & 0.99994\\
    Translation $y$ & 0.96745 & 0.95924 & 0.96496 & 0.99662 & 0.99996\\
    Translation $z$ & 0.96882 & 0.95807 & 0.96940 & 0.99708 & 0.99997\\
    Rotation $x$ & 0.78624 & 0.73519 & 0.35841 & 0.97437 & 0.99967\\
    Rotation $y$ & 0.88375 & 0.83200 & 0.53693 & 0.98556 & 0.99982\\
    Rotation $z$ & 0.89567 & 0.85831 & 0.68065 & 0.98659 & 0.99971\\
    Gripper & 0.99976 & 0.99881 & 0.99619 & 0.99963 & 0.99998\\
    \midrule
    Mean $R^2$ $\uparrow$ & 0.92151 & 0.89712 & 0.77909 & 0.99064 & 0.99986\\
    Action MSE $\downarrow$ & 0.001763 & 0.002207 & 0.002430 & 0.000211 & 0.000004\\
    Mean tokens $\bar K$ & 8 & 8 & 8 & 43.75 & 224\\
    Token efficiency $\uparrow$ & 0.11519 & 0.11214 & 0.09739 & 0.02265 & 0.00446\\
    \bottomrule
  \end{tabular*}
\end{table}

At eight tokens, \method exceeds OAT in every dimension, most strongly in rotation. FAST and Bin achieve higher mean $R^2$ with longer sequences. We define \emph{token efficiency} as $E_{\mathrm{tok}}=\overline{R^2}/\bar K$, mean $R^2$ divided by mean content-token count. FAST's $\bar K$ is averaged over validation chunks. This descriptive ratio favors shorter sequences and should be read alongside absolute quality and sequence length; it does not measure inference speed.

\section{OAT with Matched Action Coordinates}
\label{app:matched-coordinates}
To examine whether the action-coordinate representation accounts for \method's advantage over OAT, we train an additional OAT tokenizer and policy on LIBERO using the same cumulative translation coordinates as \method. Normalized translation commands are cumulatively summed within each chunk, while rotation and gripper channels retain per-step values. Decoded translations are differenced back into per-step commands before computing OAT's reconstruction loss and before execution. The coordinate transform adds no parameters; OAT's encoder and decoder architectures, quantizer, and nested-dropout scheme are retained.

Table~\ref{tab:matched-coordinates} compares all three configurations at eight tokens. The coordinate-matched OAT achieves 62.56\% success, compared with 61.48\% for the original OAT and 71.16\% for \method. The remaining 8.60-percentage-point gap indicates that matching translation coordinates alone does not account for \method's advantage in this comparison.

\begin{table}[!htbp]
  \caption{\textbf{LIBERO comparison with matched action coordinates.} Success (\%) at $K=8$, reported as mean $\pm$ SEM across five evaluation rounds. Rotation and gripper channels remain per-step in every row.}
  \label{tab:matched-coordinates}
  \centering\small
  \begin{tabular}{llc}
    \toprule
    Method & Translation coordinates & Success\\
    \midrule
    OAT & Per-step & $61.48\sem{0.92}$\\
    OAT & Cumulative & $62.56\sem{0.67}$\\
    \midrule
    \method & Cumulative & $\mathbf{71.16}\sem{0.68}$\\
    \bottomrule
  \end{tabular}
\end{table}

\end{document}